\documentclass[letterpaper, 10 pt, conference]{ieeeconf} 
\IEEEoverridecommandlockouts 
\usepackage{cite}
\usepackage{graphicx}
\usepackage{textcomp}
\usepackage{xcolor}
\usepackage{comment}
\usepackage{amssymb}
\usepackage{amsmath}
\usepackage{siunitx}
\usepackage{soul}
\usepackage{subcaption}
\usepackage{siunitx}
\usepackage[font=footnotesize]{caption}
\usepackage{hyperref}
\usepackage{mathtools}
\usepackage[ruled,vlined,linesnumbered]{algorithm2e} 

\SetCommentSty{mycommfont}
\usepackage{bm}
\usepackage{mdwmath}
\usepackage{amsmath}

\DeclareMathOperator*{\argmin}{arg\,min}

\newtheorem{defn}{Definition}[section]

\newcommand*{\final}{\textcolor{black}}
\begin{document}
%
\title{\LARGE \bf Reactive Locomotion Decision-Making and Robust Motion Planning\\ for Real-Time Perturbation Recovery}

%
%
%

\author{Zhaoyuan Gu, Nathan Boyd, and Ye Zhao
\thanks{The authors are with the Laboratory for Intelligent Decision and Autonomous Robots, Woodruff School of Mechanical Engineering, Georgia Institute of Technology. {\tt\footnotesize \{zgu78, nboyd31, yezhao\}@gatech.edu}}
\thanks{This work was funded by the NSF grant \# IIS-1924978 and Georgia Tech Institute for Robotics and Intelligent Machines Seed Grant.}
}

\markboth{Journal of \LaTeX\ Class Files,~Vol.~14, No.~8, August~2015}%
{Shell \MakeLowercase{\textit{et al.}}: Bare Demo of IEEEtran.cls for IEEE Journals}
%



\maketitle

\begin{abstract}

In this paper, we examine the problem of push recovery for bipedal robot locomotion and present a reactive decision-making and robust planning framework for locomotion resilient to external perturbations. Rejecting perturbations is an essential capability of bipedal robots and has been widely studied in the locomotion literature. However, adversarial disturbances and aggressive turning can lead to negative lateral step width (i.e., crossed-leg scenarios) with unstable motions and self-collision risks. These motion planning problems are computationally difficult and have not been explored under a hierarchically integrated task and motion planning method. We explore a planning and decision-making framework that closely ties linear-temporal-logic-based reactive synthesis with trajectory optimization incorporating the robot's full-body dynamics, kinematics, and leg collision avoidance constraints. Between the high-level discrete symbolic decision-making and the low-level continuous motion planning, behavior trees serve as a reactive interface to handle perturbations occurring at any time of the locomotion process. Our experimental results show the efficacy of our method in generating resilient recovery behaviors in response to diverse perturbations from any direction with bounded magnitudes.

\end{abstract}


%
\IEEEpeerreviewmaketitle

\section{Introduction}
As legged robots are increasingly deployed in complex environments, the need for robots to accomplish tasks through symbolic planning and decision-making becomes more apparent. Although locomotion robustness has been extensively explored at the motion planning level, resilience to uncertainties and external disturbances at the task planning level has been largely overlooked. Hierarchically integrated task and motion planning (TAMP) is capable of handling logical and whole-body dynamics objectives simultaneously. Unexpected errors or even failures at the lower-level can lead to expensive re-planning at the higher task planning level. On the other hand, high-level discrete task plans can result in infeasible low-level motion plans. With these cascading effects, novel TAMP methods are imperative to make robust locomotion decisions resilient to environmental perturbations and enable robots to efficiently recompute plans at both task and motion planning levels. 

At the motion planning level, push recovery of bipedal locomotion has been extensively studied in previous works and inspired by human locomotion biomechanics \cite{stephens2011push, MPC_perturbation}. Various strategies such as hip, ankle, and foot placement strategies are proposed to handle external perturbations \cite{pratt2006capture, yi2011online, shafiee2019online}.
However, many of these push recovery strategies employ reduced-order models (RoMs) such as inverted pendulum or centroidal momentum models, making it difficult to guarantee leg self-collision avoidance. This challenge arises from solving full-leg kinematic constraints in these RoMs. It is a strong assumption to state that a robot will never be in close contact with itself in highly dynamic locomotion. Liu et al. \cite{Chengju_et_al2017} demonstrated a complete control framework that considers self-collision under various disturbances, but does not consider more complicated multi-step or non-periodic recoveries. Reactive approaches for high dimensional robots have also been explored \cite{Zhou_reactive2016,Hildebrandt2014,Dietrich2011}, which rely on a distance metric to generate safe repulsive motions, but can lead to significant motion plan discrepancies. Behavior libraries have also been used to generate robust real-time walking in unstructured or constrained environments \cite{Nguyen2016, gong2018feedback}. In addition, few motion planning strategies incorporate higher-level task planning. 

\begin{figure}[t]
\centerline{\includegraphics[width=.5 \textwidth]{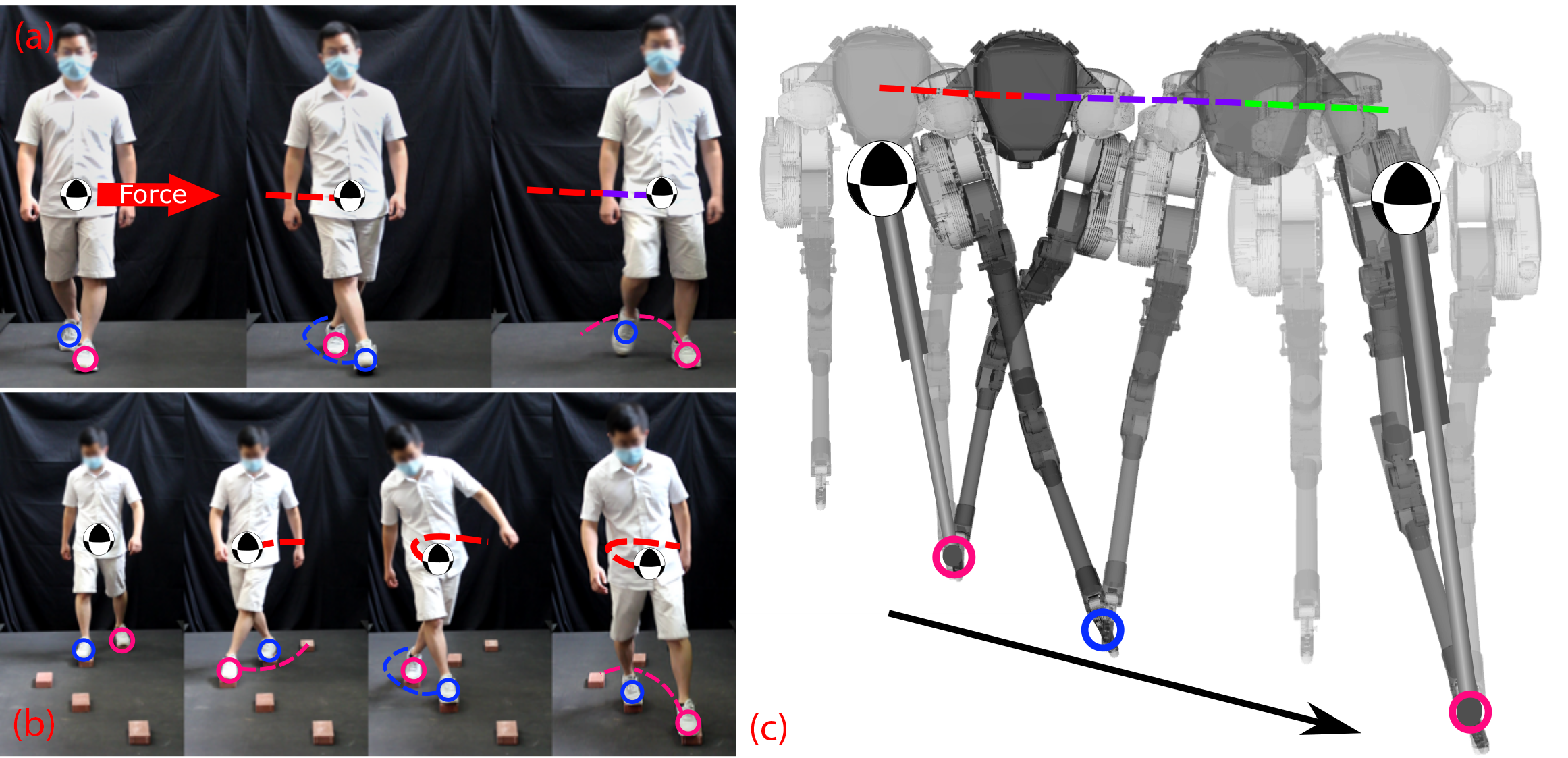}}
\caption{
a) Human is forced to cross legs to recover from an external disturbance. b) Human must execute leg crossing to traverse stepping stones. c) An illustration of recovery motion of bipedal robot Cassie.}
\label{fig:vision}
\vspace{-0.15in}
\end{figure}

For high-level task planning, reactivity is critical to account for environmental changes at runtime. Temporal-logic-based reactive synthesis  \cite{kress2009temporal, liu2013synthesis, LTL_horizon} has been widely explored to find strategies that generate formally-guaranteed safe and provably correct robot actions in response to environmental events. However, this method has been under-explored for dynamic locomotion problems until recent years. Recent works \cite{zhao2018reactive, LTL_Nav_Kulgod, LTL_Nav_Warnke, shamsah2021integrated} adopted linear temporal logic (LTL) to synthesize reactive locomotion navigation plans over rough terrains. \final{Although bipedal walking only involves alternating left and right foot contacts, incorporating external perturbations into formal foot placement decision-making in a provably correct manner remains challenging.} %
Moreover, the feasibility of executing synthesized task plans on high degree-of-freedom legged robots is unexplored. \final{To address these challenges, this study combines collision-avoidance-aware trajectory optimization (TO) with LTL methods to guarantee the task completion of the robot locomotion.} 

\begin{figure*}[t]
\centerline{\includegraphics[width=0.85\textwidth]{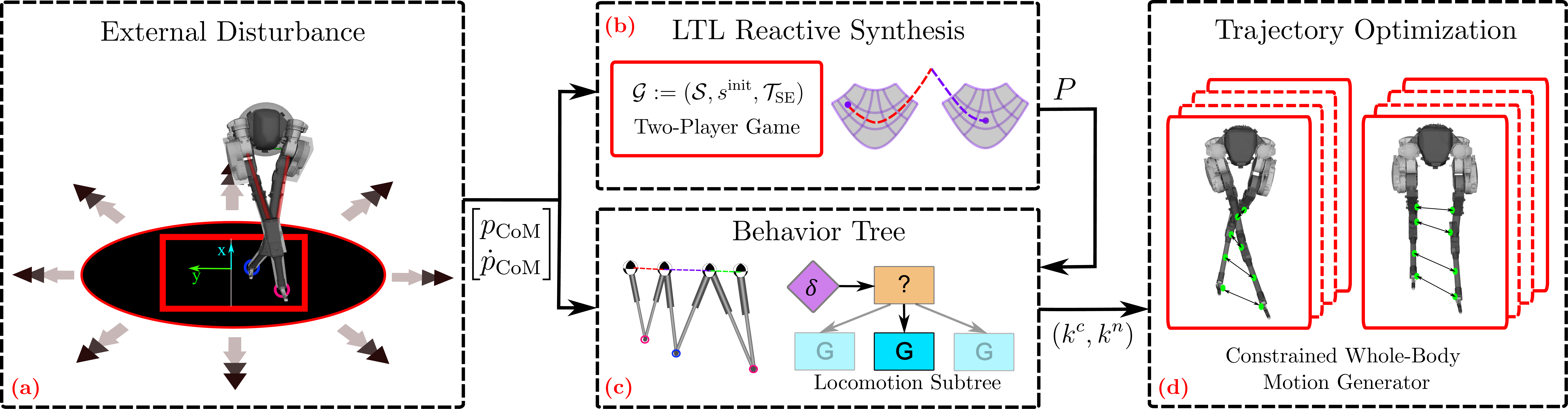}}
\caption{Block diagram of the proposed framework. a) Experiments of Cassie disturbed during stable walking; b) The high-level task planner synthesis, employing an LTL two-player game; c) The BTs act as a middle layer that reactively execute subtrees based on real-time environmental disturbances; d) A whole-body motion planner is used to generate feasible motions and refine LTL specifications $\psi$. The high-level task planner and the phase-space planner are integrated in an \textit{online} fashion as shown by the solid black arrows.}
\label{fig:framework}
\vspace{-0.15in}
\end{figure*}

Behavior Trees (BTs), as graphical mathematical models, have been widely explored to schedule autonomous tasks and handle unexpected environmental changes \cite{marzinotto2014towards, li2021reactive}. Their reactive and modular structure can authorize multiple behavioral plans and achieve fault-tolerant task executions  \cite{colledanchise2018behavior, iovino2020survey}. \cite{Park2013FSMterrain} devised finite state machine (FSM) controllers for unexpected terrain height variation, but relied on large handmade state machines. Intuitively speaking, BTs can be viewed as a feature-rich, acyclic version of FSM for complex behavior execution. Recent LTL-based reactive synthesis work \cite{zhao2018reactive} proposed reactive TAMP in combination with robust reachability analysis for dynamic maneuvers and disturbance rejection, but only accounts for perturbations applied at specific instances. Formal methods handling perturbation at any locomotion phase require further investigation. The BTs naturally handle the continuous environmental perturbations by designing actions online to amend the synthesized discrete automaton.

This study addresses the push recovery problem for legged robots subject to external perturbations that can happen anytime. We propose a combined TAMP framework composed of hierarchical planning layers operating at different temporal and spatial scales (Fig.~\ref{fig:framework}). First, the LTL planning designs safety-guaranteed decisions on keyframe states, including center of mass (CoM) state or foot placements, in response to the keyframe perturbations. When perturbations occur at non-keyframe instants, analytical Riemannian manifolds are used to recalculate a new keyframe transition online for the current walking step. BTs are integrated to allow updated keyframes to be any continuous value within the allowable range, instead of a finite set of discrete values quantified in the LTL-based planner. Finally, full-body legged motions are generated using kinodynamic-aware TO for non-periodic multi-step locomotion with self-collision constraints. \final{Compared to our previous robust locomotion work \cite{Zhao2017IJRR, zhao2016robust}, this work (i) studies perturbation recovery from comprehensive perturbations from all directions and during various locomotion phases, and (ii) solves full-body TO to generate dynamically feasible trajectories that will refine high-level decisions.}

The core contributions of this paper are summarized as:
\begin{itemize}
    \item We present a hierarchically integrated LTL-BT TAMP framework for dynamic locomotion that reacts to continuous environmental perturbations for resilient task execution.
    \item We employ Riemannian manifolds to quantify locomotion keyframe robustness margins and design robust transitions enabled by the reactive task planner.
    \item We propose a collision-aware, kinodynamic TO that generates collision-free and non-periodic full-body motions and use this TO to refine feasibility specifications in reactive synthesis.
\end{itemize}

\section{Planning Methods}
This section details the symbolic decision-making and motion planning framework (Fig.~\ref{fig:framework}). Our hierarchical reactive framework is composed of (i) LTL-level reactive synthesis handling perturbations at keyframe instants, (ii) BT for robust execution of one walking step (OWS) between keyframe instances, (iii) full-body motion primitive generation from kinodynamic-aware TO.

\subsection{Keyframe-based Non-periodic Locomotion} \label{keyframe_loco}
To define a multi-step walking motion for bipedal robot walking, we separate the entire trajectory into multiple OWS phases that start and end at keyframe states. \final{The keyframe state is defined based on a step-to-step discretization of the continuous walking process, allowing the robot to make CoM apex parameter decisions for each walking step.} The $i^{th}$ OWS cycle can be represented by a discrete keyframe transition pair $(k^i, k^{i+1})$. The keyframe contains the sagittal and lateral CoM apex state, as well as the stance foot index (Sec. \ref{LTL_spec}). Given two consecutive keyframe states in the sagittal plane, forward and backward numerical integration is used to solve for the contact switching time of a OWS ($t_1$ and $t_2$ for the first-half and second-half OWS phases, respectively). Here $t_1$ and $t_2$ are not fixed, therefore the contact switch timing is not constant and it enables non-periodic locomotion. \final{The numerical integration is based on the linear inverted pendulum dynamics.} The next sagittal keyframe will determine the next lateral keyframe state.
Namely, given two consecutive keyframe states in the sagittal plane, the next lateral keyframe can be calculated by meeting the $t_1$ and $t_2$ timing constraint, due to the simultaneous contact switch in both directions. Then the lateral keyframe transition is determined as well as the lateral CoM state $(p_{\rm switch,l},\dot{p}_{\rm switch,l})$ at contact switch instant. The next lateral foot placement can be computed with the analytical solution: 
\begin{equation}
\begin{aligned}
p_{\rm foot,l} = p_{\rm switch,l} + \frac{(e^{2\omega_{\rm asym} t_2}-1)\dot{p}_{\rm switch,l}}{(e^{2\omega_{\rm asym} t_2}+1)\omega_{\rm asym}}
\label{eq:anal_keys}
\end{aligned}
\end{equation}
where the asymptote slope $\omega_{\rm asym} = \sqrt{g/h_{\rm apex}}$ and $h_{\rm apex}$ is the relative apex CoM height with respect to the stance foot height. $g$ is the gravity constant. \final{The subscript $\rm l$ indicates the lateral space.}

Compared to periodic walking, where the robot repeats the same motion pattern periodically, keyframe-based walking allows for non-periodic walking that better accommodates rough terrain and environment disturbances. 

\subsection{LTL Specifications for Push Recovery}
\label{LTL_spec}

As the complexity of locomotion tasks increases, making safe decisions on keyframe states to recovery from push becomes intricate. To address this challenge, we employ reactive synthesis, which is built upon task specifications and an abstraction of dynamical systems \cite{kress2011correct, liu2013synthesis}. 
The tasks are represented by LTL specifications, which describe temporal and logical relations of the system properties. The abstraction (i.e., transition system) is a discrete description of the system and environment dynamics. An LTL formula operates over atomic propositions (APs) that can be \textsf{True} ($\varphi \vee \neg\varphi$) or \textsf{False} ($\neg \textsf{True}$). The formulas use logical symbols of negation ($\neg$), disjunction ($\vee$), and conjunction ($\wedge$). Temporal operators such as ``next" ($\bigcirc$), ``eventually" ($\Diamond$), and ``always" ($\square$) are used as extensions to the propositional logic. Detailed LTL semantics are omitted due to space limit and can be found in \cite{baier2008principles}.

To formally guarantee locomotion task completion under environmental disturbances, we adopt the General Reactivity of Rank 1 (GR(1)) \cite{gr1}, a fragment of LTL. GR(1) provides correct-by-construction guarantees of the realizability of LTL specifications.
Provided a transition system $\mathcal{T}_{\rm SE}$ and LTL specification $\psi$, the reactive synthesis problem aims for a winning strategy for the robot system such that the execution path satisfies $\psi$ \cite{LTL_Nav_Warnke}. If the specification is realizable, an automaton will be constructed and provide correct transitions for any environmental actions obeying the assumptions. 

\begin{figure}[t]
\centerline{\includegraphics[width=.4\textwidth]{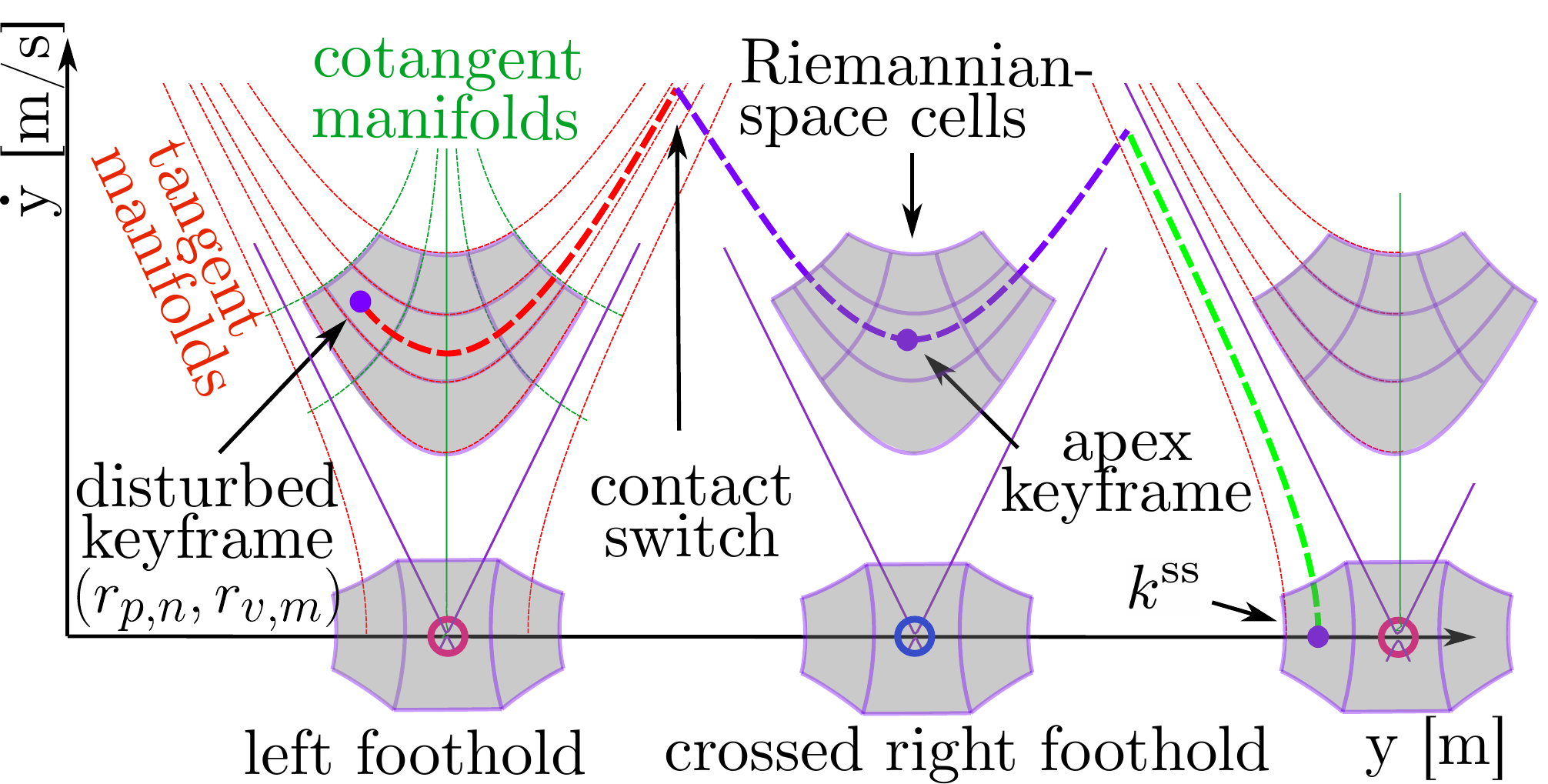}}
\caption{
An illustration of a phase-space Riemannian partition and non-deterministic lateral keyframe transition for disturbance recovery.}
\label{fig:riem}
\vspace{-0.15in}
\end{figure}

\begin{defn}[Riemannian partition]\label{def:riem}
The transition system discretizes the continuous robot state space (i.e., robot's CoM phase-space near the apex state) into Riemannian partitions defined as:
$$
\begin{aligned}
\mathcal{R} &\coloneqq \mathcal{R}_{\rm position} \final{\times} \mathcal{R}_{\rm velocity} \\
&\,= \{r_{p,n}, r_{p,z}, r_{p,p}\} \final{\times} \{r_{v,z}, r_{v,s}, r_{v,m}, r_{v,f}\}
\end{aligned}
$$
where the elements in $\mathcal{R}_{\rm position}$ define the relative position (negative, zero, positive) of CoM with respect to the stance foot frame, and $\mathcal{R}_{\rm velocity}$ defines the CoM apex velocity (zero, slow, medium, fast). Riemannian partitions are defined for both sagittal and lateral phase-space, each constitutes 12 cells. 
\end{defn}

Fig.~\ref{fig:riem} shows a disturbed keyframe state $(r_{p,n}, r_{v,m})$, which represents a negative position and medium velocity. A keyframe state whose CoM velocity is zero in sagittal axis is noted as $(r_{v})_s = (r_{v,z})_s$. The Riemannian partitions use the analytical manifolds of CoM dynamics derived from the Prismatic Inverted Pendulum Model (PIPM). More details will be introduced in Sec.~\ref{riem}.

\begin{defn}[Locomotion keyframe]\label{def:keyframe}
A keyframe $\mathcal{K}$ is defined as a system apex state composed of the sagittal partition $\mathcal{R}_{s}$, the lateral partition $\mathcal{R}_{l}$, as well as the stance foot index set $\mathcal{F}_{\rm st} = \{\textsf{left}, \textsf{right}\}$ (used to identify the leg crossing or wider lateral step strategies). 
$$\mathcal{K} \coloneqq \mathcal{R}_{s} \final{\times} \mathcal{R}_{l} \final{\times} \mathcal{F}_{\rm st}.$$
\end{defn}

The system takes actions $a_{\rm sys} \in \mathcal{A}_{\rm sys} \subseteq \mathcal{R}_{s} \final{\times} \mathcal{R}_{l} \final{\times} \mathcal{L} \final{\times} \mathcal{W}$ to decide the next keyframe state $k^n$. 
$\mathcal{L} = \{\textsf{small}, \textsf{medium}, \textsf{large}\}$ and $\mathcal{W} = \{\textsf{small}, \textsf{medium}, \textsf{large}\}$ represent the step length and width. $l \in \mathcal{L}$ and $w \in \mathcal{W}$ are the \final{nominal} distances between the current and the next foot placements projected on sagittal and lateral axis. Note that $\mathcal{L}, \mathcal{W}$ are the \final{nominal} global distances between footholds while $\mathcal{R}_{s}, \mathcal{R}_{l}$ are the relative CoM apex states in \final{nominal} foot frame. 

The environment state is represented by a perturbation set $p_{\rm env} \in \mathcal{P}_{\rm env} \coloneqq \mathcal{R}_{s} \final{\times} \mathcal{R}_{l} \cup \{ \emptyset\}$ that pushes the system to a specific Riemannian cell center. In the task planner, we assume that the environment action is a perturbation only applied at a keyframe instant. The perturbation induces a CoM position and velocity jump after applying an external force to the robot's pelvis frame. The environment can also choose to not perturb, i.e., $p_{\rm env} = \emptyset$. The system action $\mathcal{A}_{\rm sys}$ and environment action $\mathcal{P}_{\rm env}$ together decide the next apex keyframe state $k^{n} = \mathcal{T}_{\rm SE}(k^c, a_{\rm sys}, p_{\rm env})$. Both actions are a part of the automaton state $\mathcal{S}$.

\begin{defn}[Steady state keyframe]\label{def:ss}
A special set of keyframes are defined as steady state keyframes $k^{\rm ss} \in \mathcal{K}^{\rm ss}$ during perturbation-free walking.
$$\mathcal{K}^{\rm ss} = \{k^{\rm ss} | k^{\rm ss} = \big((r_{p,z}, r_{v, \cdot})_s, (r_{p,\cdot}, r_{v,z})_l, f_{\rm st}\big)\}$$ 
where $(r_{p,z}, r_{v, \cdot})_s$ means that the sagittal CoM apex position is on top of the nominal foot placement and can take any allowable sagittal velocities, while $(r_{p,\cdot}, r_{v,z})_l$ means that the lateral CoM apex position can take any values and the apex velocity has to be zero $r_{v,z}$ (see Fig.~\ref{fig:riem}). 
\end{defn}


Let the system start from a steady state $k^{\rm ss}_{\rm sys} = ((r_{p,z}, r_{v,m})_s, (r_{p,z}, r_{v,z})_l, \textsf{right})$. We have
$$
\begin{aligned}
s^{\rm init} &= (k^{\rm init},a_{\rm sys}^{\rm init},p_{\rm env}^{\rm init}) 
\\&= \big(k^{\rm ss}, ((r_{p,z}, r_{v,m})_s, (r_{p,z}, r_{v,z})_l),\emptyset\big)
\end{aligned}
$$

The robot chooses to maintain stable walking so long as there is no perturbation from the environment. In the presence of perturbations, the keyframe state returns to a steady state within two steps: 
$$\square \big(k = \neg k^{\rm ss} \Rightarrow (\bigcirc k = k^{\rm ss})\vee(\bigcirc \bigcirc k = k^{\rm ss})\big)$$
The feasibility of transitions must be verified by the low-level full-body TO (Sec.~\ref{TO_implement}). Certain high-level transitions should be removed due to infeasible full-body kinematics and dynamics constraints. In this way, we define a set of TO-refined task specifications. For example, after the TO refinement, we obtain all full-body-dynamics-feasible transitions offline and encode TO-refined specifications. An example of TO-refined specification can be:
$$
\begin{aligned}
\square \big(k&=((r_{p,z},r_{v,m})_s,(r_{p,z},r_{v,m})_l, \textsf{right}) \Rightarrow \\ a&=((r_{p,z},r_{v,m})_s,(r_{p,z},r_{v,s})_l, \textsf{small}, \textsf{small})\\
&...\\
\vee a&=((r_{p,z},r_{v,m})_s,(r_{p,z},r_{v,m})_l,\textsf{small}, \textsf{medium}) \big)
\end{aligned}
$$
\vspace{-0.1in}

In the presence of perturbations, recovering to a steady state $k^{\rm ss}$ requires the next keyframe $k^n$ to decrease the lateral apex velocity and minimizes the sagittal apex deviation from its normal value. For example, a current \textsf{medium} apex velocity indicates the next apex velocity is either \textsf{medium}, \textsf{small} or \textsf{zero}: $\square\big(r_v=r_{v,m} \Rightarrow (\bigcirc r_v = r_{v,m} \vee r_{v,s} \vee r_{v,z})\big)$. A smaller step width $w \in \mathcal{W}$ will be chosen rather than larger ones. $\square\big((w=\textsf{small} \vee w=\textsf{large}) \Rightarrow (\bigcirc w=\textsf{small})\big)$. 

For the recovery motion execution not to be interrupted, we assume the environment perturbation happens at most once per two steps: $\square \big(p_{\rm env} = \neg\emptyset \Rightarrow (\bigcirc p_{\rm env} = \emptyset)\big)$

\subsection{Task Planner Synthesis}
\label{task_planner_synth}
Given the LTL specifications above, the task planner models the robot system and the environment interplay as a two-player game. We construct the keyframe transition game structure in the form of a tuple $\mathcal{G} := (\mathcal{S}, s^{\rm init}, \mathcal{T}_{\rm SE})$ with:
\begin{itemize}
    \item $\mathcal{S} = \mathcal{K} \times \mathcal{A}_{\rm sys} \times \mathcal{P}_{\rm env}$ is the possible automaton state of the transition system,
    \item $s^{\rm init} = (k^{\rm init},a_{\rm sys}^{\rm init},p_{\rm env}^{\rm init}) $ is the initial automaton state and
    \item $\mathcal{T}_{\rm SE} \subseteq \mathcal{S} \times \mathcal{S}$ is a transition describing the possible moves of the robot system and antagonist environment.
\end{itemize}

In the extreme case where the disturbance is towards the stance leg (see Fig.~\ref{fig:vision}), the foot placement of the swing leg would naturally move closer to the stance foothold location or require a crossed-leg motion. A minimum of two steps is required to recover in such case, during which self-collision poses a challenge for making safe decisions. The TO-refined transition specifications guarantee that the task planner makes dynamics-informed decisions on keyframe transitions. By construction, the TO indicates that all the constraints on the full-body motion are fulfilled and a keyframe transition is feasible.

At each keyframe instant, the decision maker uses the estimated current system keyframe state $k^c$ and plans a sequence of transitions until the final state $k^f = k^{\rm ss}$. The action roll-out produces an action plan $P = \{k^c, \dots, k^f\}$.


\begin{figure}[t]
\centerline{\includegraphics[width=.42\textwidth]{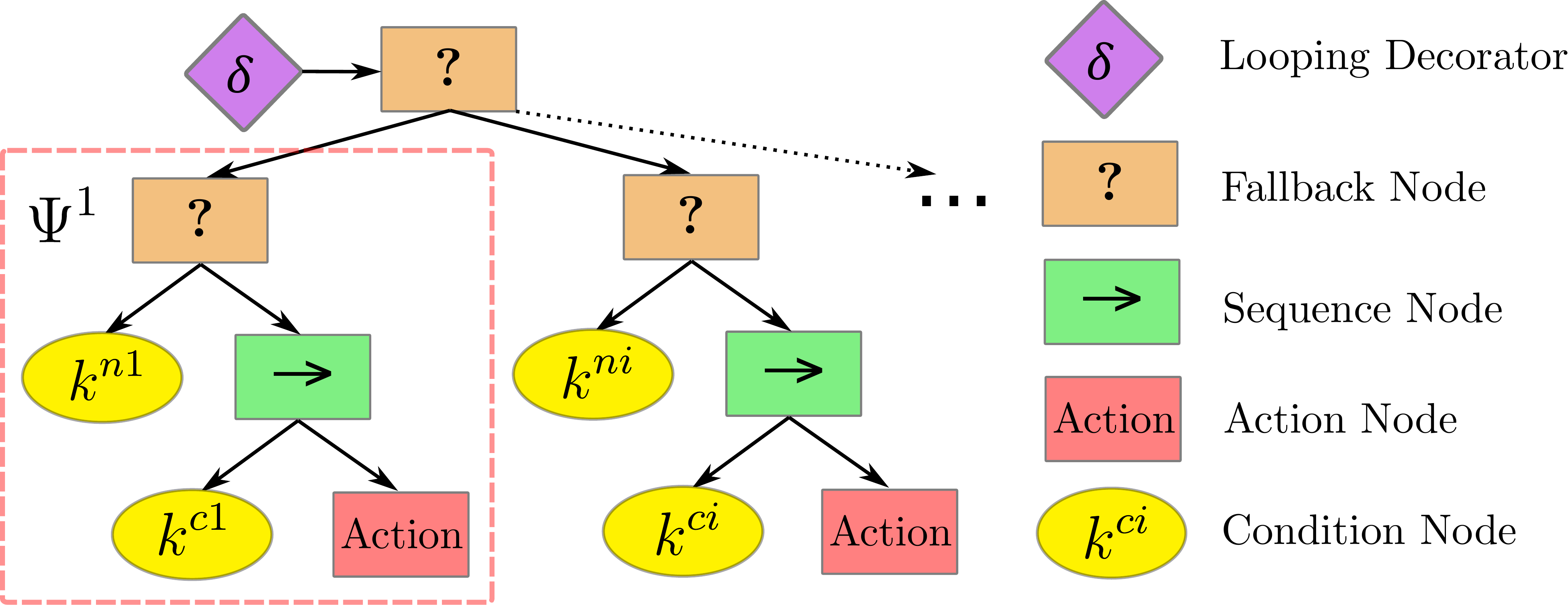}}
\caption{
An illustration of the PABT structure. The PABT groups a set of locomotion subtrees $\Psi^i$. Each subtree is a fallback tree that encodes a keyframe transition $(k^{c,i}, k^{n,i})$ and a Riemannian recalculation action.}
\label{fig:bt}
\vspace{-0.15in}
\end{figure}

\subsection{Behavior-Tree-Based Dynamic Replanning} \label{BT_implement}

To address continuous perturbations at non-keyframe instants, we propose a perturbation-aware behavior tree (PABT) that online modifies the desired keyframe transition $(k^{c,d}, k^{n,d})$. The PABT complements the reactive synthesis by locally modifying the keyframe transitions, given the real-time captured CoM state $(p_{\rm CoM}, \dot{p}_{\rm CoM})$.

The PABT groups a set of locomotion subtrees $\Psi = \bigcup\limits_{i}\Psi^i$. Each $\Psi^i$ encodes a pair of the current-to-next keyframe states $(k^{c,i}, k^{n,i})$. These pairs are represented as condition nodes in the locomotion subtrees (Fig.~\ref{fig:bt}). 
The locomotion subtrees are fallback BTs that execute their action nodes when the desired keyframe transition from the high-level matches their condition nodes. For instance, the pre-condition nodes check if the desired transition $k^{c,d}$ matches with their keyframe condition $k^{c,i}$, the same for the post-condition nodes.

The PABT modifies its keyframe transitions locally to handle non-keyframe perturbations. After the modification, the desired keyframe transition remains feasible despite the CoM state deviation. The action node $A^i$ can also be a keyframe recalculation procedure. Here we use the recovery strategy \cite{Zhao2017IJRR} to perform a Riemannian recalculation, which recalculate the keyframe transition when the CoM state is perturbed off from the nominal manifold. It is worth noting that the PABT modified keyframe state may not be a Riemannian cell center or even end up in a different cell. 

\setlength{\textfloatsep}{0.5cm}
\setlength{\floatsep}{0.5cm}
\begin{algorithm}[t]
\SetAlgoLined
\SetNoFillComment
\textbf{Input:} PABT $\Psi$, Decision Maker $DM$, current $time$\;
\textbf{Set:} status = \textsf{success}\;
\While{ status == \textnormal{\textsf{success}} }{
    $k^c, time$ = StateEstimation()\;
    \If{ $time$ == \textnormal{\textsf{keyframe\_instant}}} {
        $P = DM(k^c)$\;
        \For {$(k^c, k^n)$ \textnormal{in} $P$} {
            $\Psi^c$ = LocomotionSubtree$(k^c, k^n)$\;
            $\Psi$.Insert($\Psi^c$)\;
        }
    }
    \tcc{PABT Riemannian Recalculation}
    $status$ = $\Psi$.Tick()\;
    $(k^{c}, k^{n})'$ = $\Psi$.GetModifiedTransition()\; 
}
\textbf{Output:} updated PABT $\Psi$, modified keyframe transition $(k^{c}, k^{n})'$\;
\caption{Keyframe Decision Making and PABT Execution}
\label{alg:LTL_BT}
\end{algorithm}

The PABT grows as the new action plan $P$ is commanded from the task planner. The PABT constructs new subtrees $\Psi^c$ that represent the transitions $(k^c, k^n)$ from $P$. The new subtrees are inserted under the root node as new behaviors. A tick of the PABT will trigger the corresponding subtree that matches the subtree conditions. 
The PABT expansion and execution process is illustrated in \final{Algorithm}~\ref{alg:LTL_BT}.

\subsection{Riemannian Robustness Margin Design}
\label{riem}

To quantify robustness margin, we use the Riemannian distance metric to measure the deviation of CoM state from the nominal CoM manifolds in the CoM phase-space. This Riemannian metric discretizes the phase-space with tangent and cotangent locomotion manifolds, instead of using na\"ive Euclidean-type discretization. The tangent and cotangent manifolds comply with the PIPM locomotion dynamics and provide an intuitive trajectory recalculation strategy for CoM deviation.


We use the position guard strategy \cite{Zhao2017IJRR} to recalculate the next CoM apex state. Assuming the CoM state jumps to $(p_{\rm CoM}', \dot{p}_{\rm CoM}')$ on a new tangent manifold $\sigma'$, the recalculated next CoM apex state is:
\begin{equation}
\begin{aligned}
(p_{\rm apex}, \dot{p}_{\rm apex}) = (p_{\rm foot},\sqrt{\frac{\dot{p}_{\rm CoM}'^2 \pm \sqrt{\dot{p}_{\rm CoM}'^4-4\omega_{\rm asym}^2\sigma'}}{2}})
\end{aligned}
\end{equation}
Note that the $(p_{\rm apex}, \dot{p}_{\rm apex})$ corresponds to the next keyframe $k^n$ at the LTL level. The motion primitive set interpolates a full-body motion that connects the current CoM state to the updated next keyframe.

\subsection{Collision-Aware Kinodynamic Trajectory Optimization}
\label{TO_implement}
The task planner and PABTs generate keyframe transitions robust to perturbations. However, mapping the transitions to whole-body trajectories in real-time often poses a challenge due to the curse of dimensionality. To address this, we use TO to create a set of motion primitives offline. The TO generates desired motions that satisfy the physical constraints while minimizing the trajectory cost \cite{RaoOptimalControl, Frost, RT_collision}. The TO is also used as a verification to check the feasibility of high-level keyframe transitions.
The nonlinear program (NLP) of TO is formulated as:
%
\begin{align}
\label{eq:optimization}
\argmin_{X} \quad  & \sum_{j=1}^{D}  \sum_{i=0}^{N_j} \Omega_j \cdot \mathcal{L}_j(x^j_i, u^j_i) \\\nonumber
\textrm{s.t.} \quad & H_j(x^j_i) \dot{x}_i + V_j(x^j_i) + G_j(x^j_i) = u_i, \quad (\textnormal{dynamics})  \\\nonumber
                & x_{0}^{j+1} = \Delta_j(x_{N_j}^j),  \qquad\qquad\qquad\qquad\,\, (\textnormal{reset\,map}) \\\nonumber
                 & \lambda_{c,z}\geq 0,  \; |\lambda_{c,xy}| \leq \mu \lambda_{c,z}, \qquad\qquad\quad\;\;\,  (\textnormal{friction}) \\\nonumber
                & C_j^{\rm kin}(x^j_i) \leq 0, \qquad\qquad\qquad\qquad\quad\,\,(\textnormal{kinematics})  \\\nonumber
                & C_j^{\rm col}(x^j_i) \leq 0,  \qquad\qquad\quad\qquad\;\;\;\;\;\,(\textnormal{self-collision}) \\\nonumber
                & C_j^{\rm key}(x^j_i, u^j_i) = 0  \qquad\qquad\;\;\; (\textnormal{keyframe\;boundary})
\end{align}
\noindent where the domains include $D=2$ continuous single stance phases and one velocity reset map. Each single stance contains $N_j$ nodes, which represents a state-control pair $n^j_i=(x^j_i, u^j_i)$ at the $i^{\rm th}$ instant; the state represents the $x = [q;\dot{q}]$ with $q$ denoting the robot generalized coordinate states.
The NLP above solves the optimal state-control trajectory $X^* = \{n^{j*}_i\}$ by minimizing the pseudo energy $\mathcal{L}(\cdot) = || u_i ||^2$ with weights $\Omega_j$ while enforcing the physical constraints of the robot. 

The physical constraints shape the resultant trajectory.
The dynamics constraint is enforced between node points using Hermite-Simpson collocation. $H$, $V$, and $G$ denote the inertia, coriolis, and gravity matrices of the robot's rigid body dynamics.
\final{We ignore the double stance phase and model the discrete jump at the ground impact instant with $x_{0}^{j+1} = \Delta_j(x_{N_j}^j)$, which only maps a discrete jump of the velocity component for the state vector \cite{Frost}.} The horizontal contact forces $\lambda_{c,xy}$ are bounded by a linearized friction cone. 
The kinematics constraints $C_j^{\rm kin}(x^j_i)$ ensure that the joint angles, foot positions, and CoM trajectories are bounded.
$M$ geometric point pairs $(\bm{g}^m_l, \bm{g}^m_r)$ on two legs are selected as self-collision constraints (see Fig.~\ref{fig:framework}d). The signed distances are evaluated at each geometric point pair using forward kinematics $FK_{\bm{g}^m}(x_i)$ for all $m \in M$, $i \in N_j$. The minimally allowed distance for pair $m$ is denoted as $\bm{d}^m_{\rm min}$.
\begin{equation}
\begin{aligned}
\bm{d}^m(x_i) &= FK_{\bm{g}^m_l}(x_i) - FK_{\bm{g}^m_r}(x_i), \\
C_j^{\rm col}(x_i) &= ||\bm{d}^m_{\rm min}||_2^2 - ||\bm{d}^m(x_i)||_2^2.
\label{eq:col_constr}
\end{aligned}
\end{equation}
The keyframe transition $(k^c, k^n)$ commanded from the task planner is enforced as boundary conditions for the foot placement, the apex CoM position and velocity. 




\begin{figure}[t]
\centerline{\includegraphics[width=.48\textwidth]{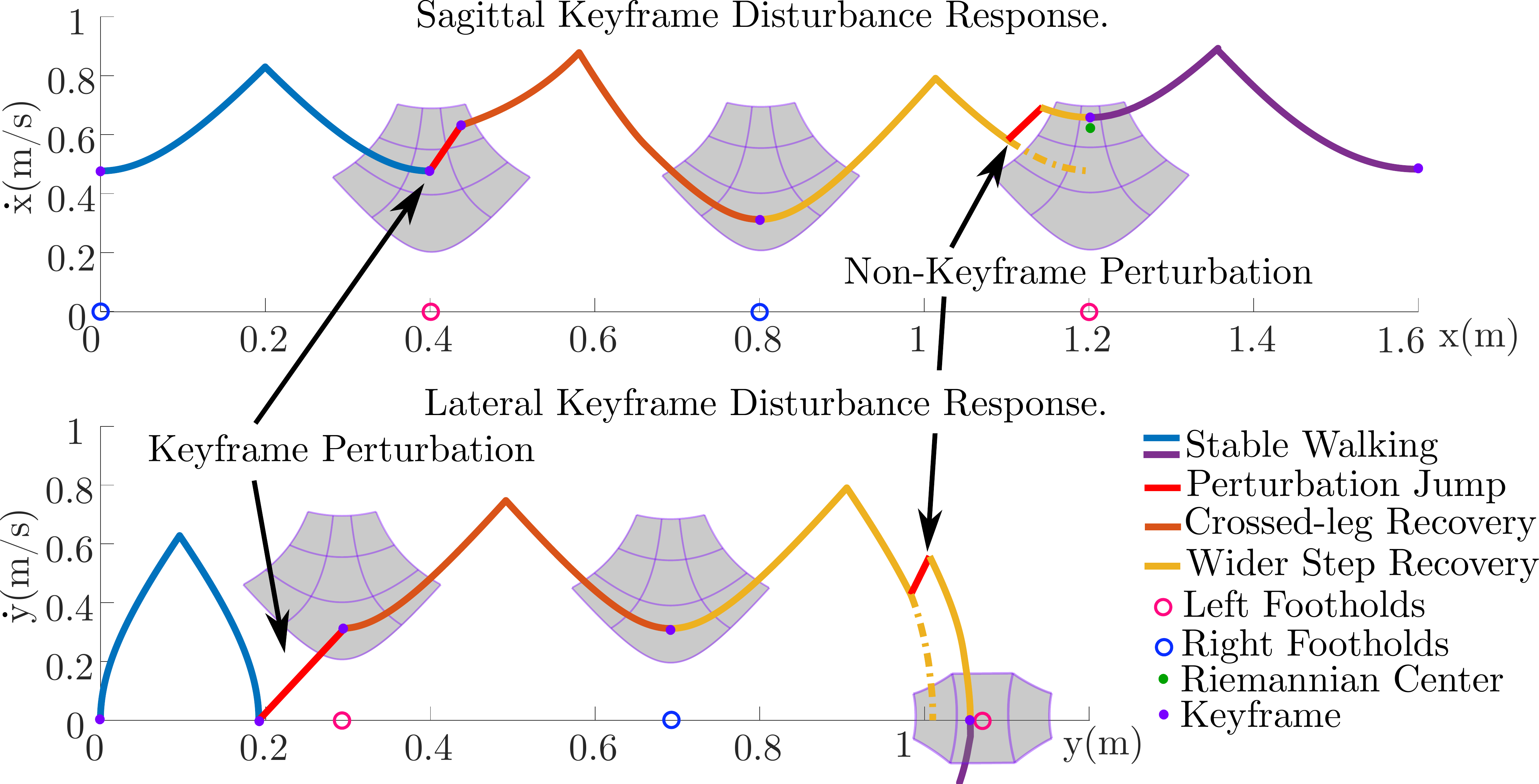}}
\caption{
Lateral and sagittal responses to diagonal disturbances at keyframe and non-keyframe instants while walking at $0.5$ m/s apex velocity. Each color represents a single step generated by the LTL-BT. }
\label{fig:ps_result}
\vspace{-0.15in}
\end{figure}

\section{Results}
To demonstrate the robustness of the proposed methods, we tested various scenarios in simulation using Matlab Simulink with a bipedal robot, Cassie. The kinematics and dynamics functions were generated using the Fast Robot Optimization and Simulation Toolkit (FROST)\cite{Frost}. The NLP solver IPOPT\cite{IPOPT} solved the TO problems \final{(\ref{eq:optimization})}. Our framework, together with a virtual constraint controller\cite{gong2018feedback}, ran at a rate of 2kHz online. Impulse forces were measured through discontinuous changes in CoM velocity. We used SLUGS reactive synthesis toolbox\cite{slugs} to design LTL specifications and synthesized the keyframe-based automaton.

For our crossed-leg experimentation, we used the 9 partitions with non-zero apex velocities for $\mathcal{R}^c_s$, $\mathcal{R}^c_l$ and $\mathcal{R}^n_s$, respectively. For each $(r^c_s, r^c_l, r^n_s)$ pair, phase-space planning calculated the next $r^n_l$. This Riemannian abstraction provided $9\times9\times9=729$ possible crossed-leg transitions prior to the full-body TO. We evaluated the feasible transitions and generated feasibility specifications. These specifications encoded the feasible high-level keyframe transitions. For stable walking and wider step recovery scenario, the lateral $r^s_l$ and $r^n_l$ were the bottom three Riemannian partitions.

\begin{figure}[t]
\centerline{\includegraphics[width=.4\textwidth]{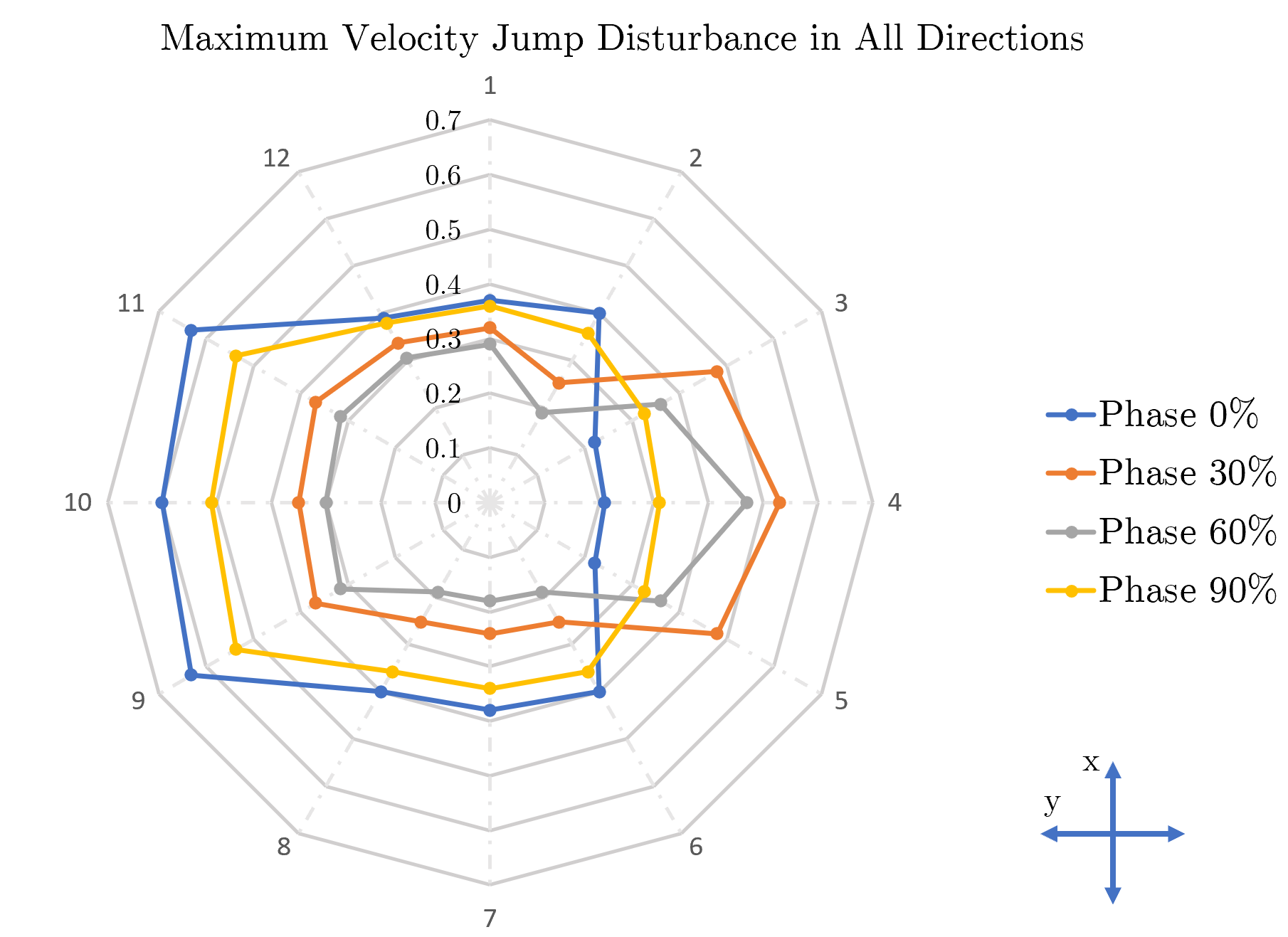}}
\caption{
Maximum allowable velocity change exerted on the CoM for a single step at $30^\circ$ increments. \final{The perturbation happens at different phases during a right leg stance.} Values on the left half resulted in single wider step recoveries and values on the right half require crossed-leg maneuvers.}
\label{fig:spider}
\vspace{-0.15in}
\end{figure}

We evaluated the performance of our framework through multiple push recovery studies. 
As shown in Fig.~\ref{fig:ps_result}, the system was capable of composing multiple OWS trajectories according to the reactive synthesis plan. The robot was firstly disturbed to the non-apex velocity $(\dot{x}, \dot{y}) = (0.63, 0.31)$ m/s at keyframe instant. The keyframe decision maker planned a two-step recovery strategy (one crossed-leg step and one succeeding wider step) to come back to a steady state $k^{\rm ss}$. Disturbances at non-keyframe states required the robot to recalculate a new CoM trajectory to an updated keyframe state. The PABT locally modified the desired keyframe transition and allowed the transitions to start and terminate in non-Riemannian-cell-centers. The reactive synthesis could update the keyframe transitions as long as the CoM state was inside the Riemannian robustness bound (grey areas in Fig.~\ref{fig:ps_result}). This preserved the notion of continuous recovery rather than \final{that of} a finite set of discrete keyframe transitions. 

\begin{figure}[t]
\centerline{\includegraphics[width=.42\textwidth]{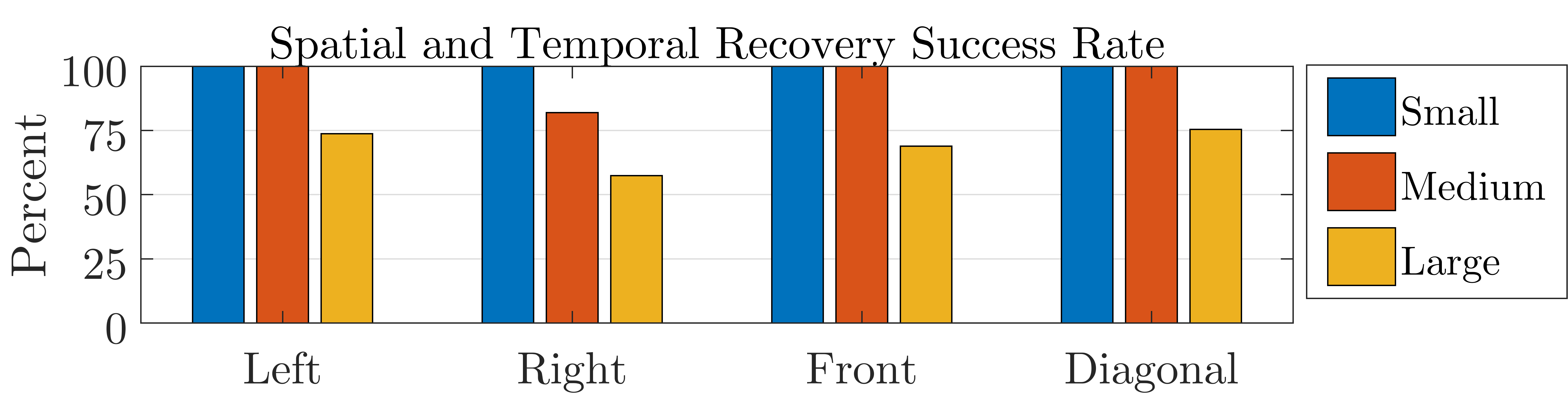}}
\caption{
Success rate of recovery motion when disturbance happens anytime during OWS at multiple directions. Three disturbances are used with a) small 0.1 m/s, b) medium 0.2 m/s, and c) large 0.3 m/s disturbances.}
\label{fig:success_rate}
\vspace{-0.15in}
\end{figure}

\begin{figure}[t]
\begin{subfigure}[b]{.31\textwidth}
\includegraphics[width=\textwidth]{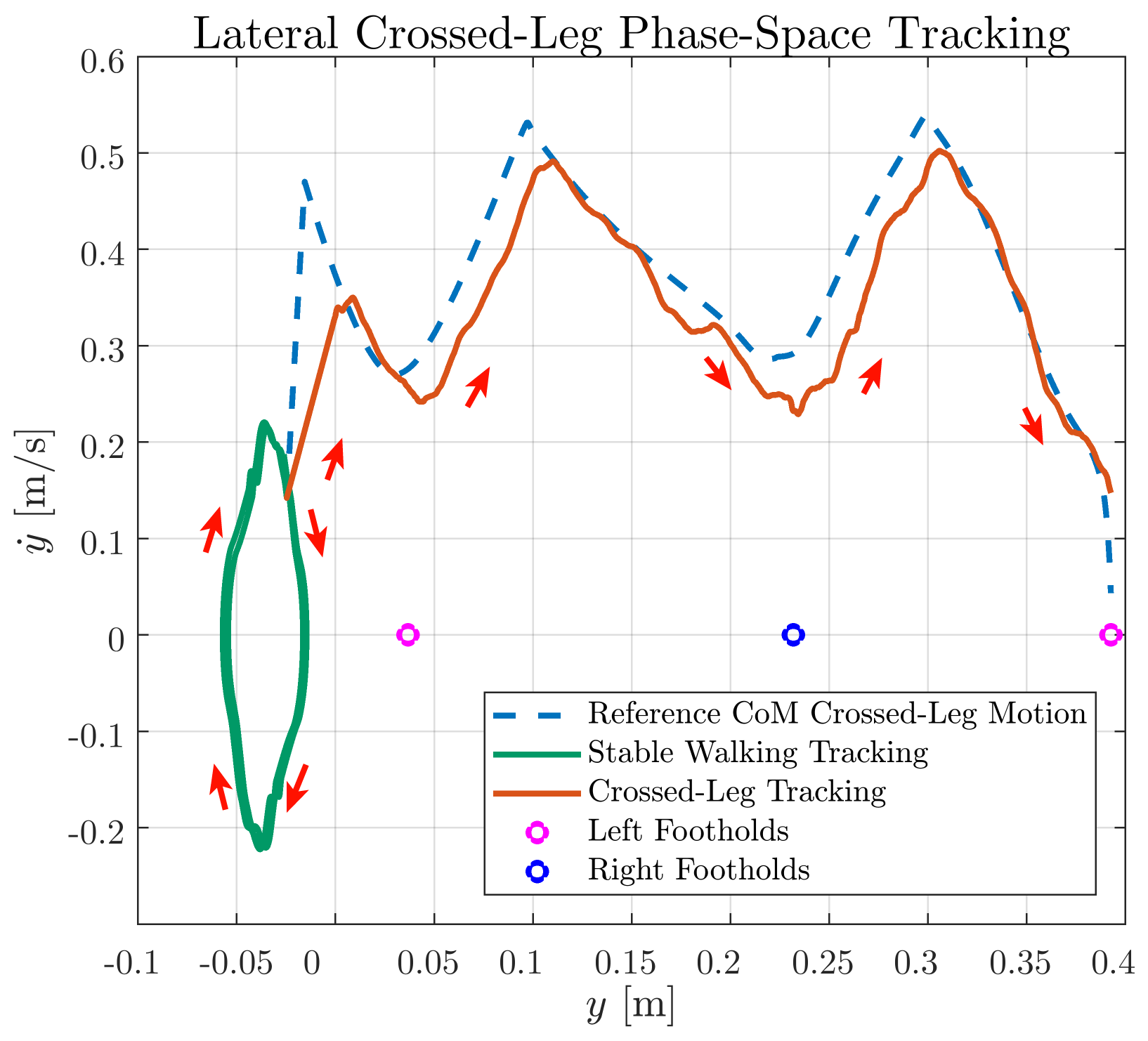}
\end{subfigure}
\begin{subfigure}[b]{.17\textwidth}
\includegraphics[width=\textwidth]{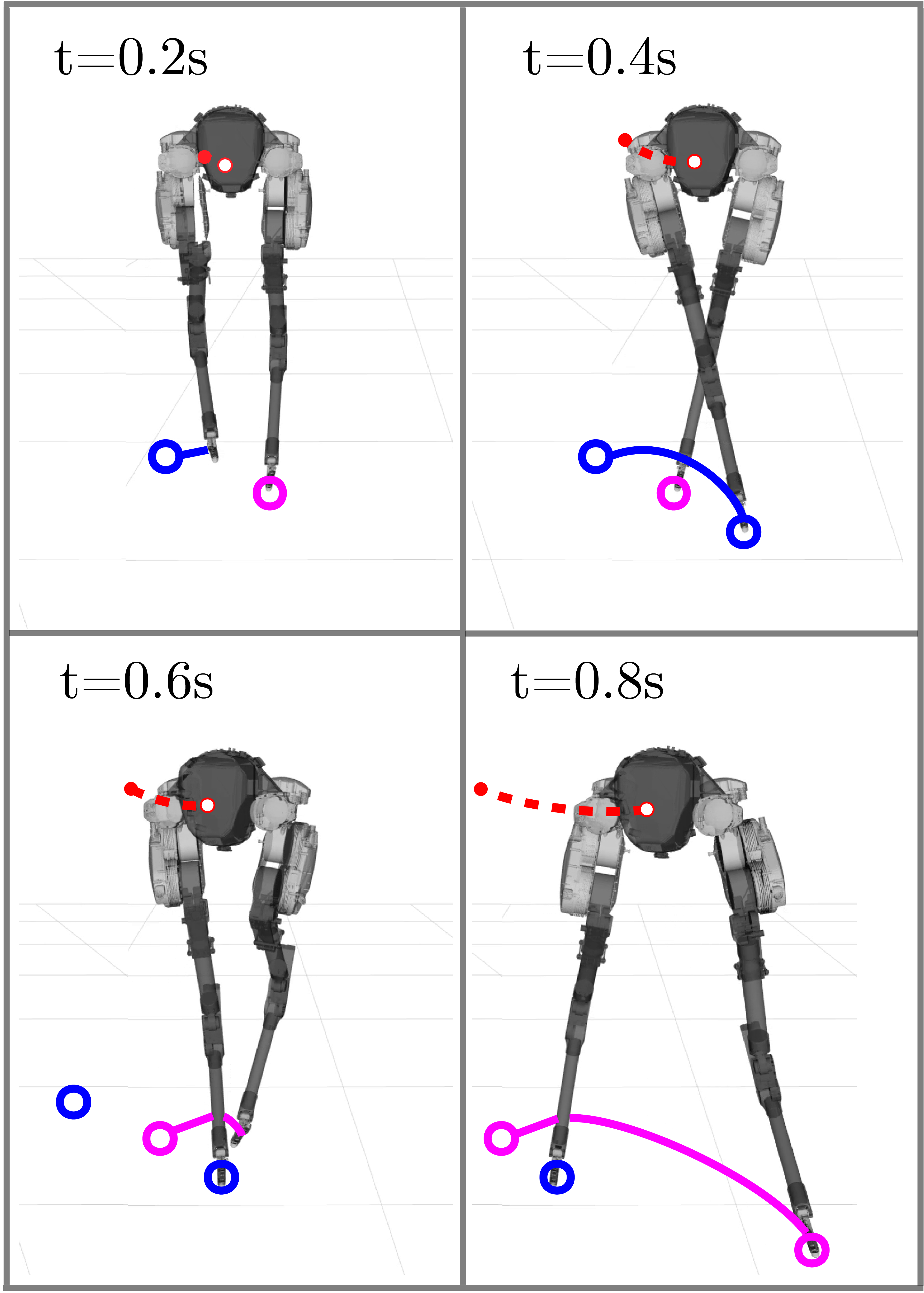}
\vspace{0.04in}
\end{subfigure}
\caption{
\final{Cassie is given a $0.4$ m/s laterally disturbance from a $0.5$ m/s stable forward walking. Cassie successfully executes a leg crossing maneuver.}}
\label{fig:tracking_result}
\vspace{-0.15in}
\end{figure}

In Fig.~\ref{fig:spider}, we compared the maximum impulse velocity changes the system can recover from in 12 directions during OWS. 
The robot walked sagittally (positive x direction) at $0.5$ m/s apex velocity. After the perturbation, it was allowed to recover using up to two steps. When the push direction was lateral left (positive y), the robot would take a wider step to come back at $k^{\rm ss}$; otherwise, when the push direction was lateral right, the robot needs to adopt the crossed-leg maneuvers. The perturbations are applied at 4 different phases, with phase $\phi=0\%$ and $90\%$ closer to keyframe states (boundary phases), and  $\phi=30\%$ and $60\%$ closer to the contact switch phase ($50\%$). The result shows that the phases close to keyframes were better at absorbing large left perturbations. Closer to the contact switch phase, the right side push is handled better due to the increased lateral velocity halfway through the step. \final{The asymmetry of the maximum allowable disturbances in the lateral directions can be attributed to the more constrained kinematic workspace of the swing legs in the crossed-leg scenario.}

We conducted an experiment to study the recovery success rate with 100 trials in 4 directions (Fig.~\ref{fig:success_rate}). Diagonal disturbances were applied at $45^\circ$ from the front to the right. For each trial, the robot took the same was disturbed with 3 instantaneous velocity jumps of $0.1,0.2,0.3$ m/s. The perturbations for each trial were spaced evenly ($\phi=1\%$) for the entire phase duration. Failures primarily occurred at the point of maximum velocity for the stance phase (right stance: $\phi\leq10\%$ and $\phi\geq90\%$, left stance: $40\%\leq \phi \leq60\%$). Similar trends 
were seen in the maximum velocity disturbances Fig.~\ref{fig:spider}. 


Finally, the tracking performance for the system was evaluated for $\pm0.4$ m/s lateral disturbances while the left leg was in stance, during a stable walking with $0.5$ m/s apex velocity. The positive disturbance forced a two-step crossed-leg recovery (Fig.~\ref{fig:tracking_result}) and had a RMS tracking error of $0.0084$ m and $0.0593$ m/s. For negative lateral disturbances, the system stabilized within one wide step with a RMS tracking error of $0.0039$ m and $0.0363$ m/s in lateral phase-space.




\section{Conclusion}
In this paper, we presented a locomotion framework for reactive disturbance rejection at the symbolic decision-making and continuous motion planning level. We combined reactive synthesis with BTs to demonstrate safe, continuous, disturbance rejection capabilities. 
At the low level, the TO generates full-body locomotion trajectories and refines feasible keyframe specifications in the reactive synthesis to fill the gap between the high-level decisions making and the low-level full-body motion planning. 
\bibliographystyle{ieeetr}
\bibliography{references}




\end{document}